%% file: main.tex
\documentclass[twoside]{article}

\usepackage{aistats2020}
\usepackage[pdftex]{graphicx}
\usepackage{wrapfig}

\usepackage{multirow}
\usepackage{appendix}
\usepackage[utf8]{inputenc} 
\usepackage[T1]{fontenc}    
\usepackage{booktabs}       
\usepackage{amsfonts}       
\usepackage{nicefrac}       
\usepackage{microtype}      
\usepackage[pdftex]{graphicx}
\usepackage{amssymb}
\usepackage{xcolor}

\usepackage{enumitem}
\usepackage{dsfont}
\usepackage{wrapfig}
\usepackage{subcaption}
\usepackage{calrsfs}

\usepackage{mathtools}
\DeclareMathAlphabet{\pazocal}{OMS}{zplm}{m}{n}

\usepackage{hyperref}
\usepackage{url}

\newcommand{\commonF}{common feature}

\newcommand{\CF}{\texttt{CF}}

\newcommand{\MCS}{\textit{MCS}}
\newcommand{\IDR}{\textit{IDR}}
\newcommand{\IIR}{\textit{IIR}}

\usepackage[round]{natbib}

\begin{document}

%

%

\twocolumn[

\aistatstitle{Benchmarking Attribution Methods with \\ Relative Feature Importance}

\aistatsauthor{ Mengjiao Yang \And Been Kim }

\aistatsaddress{ Google Brain \\ \texttt{sherryy@google.com}
  \And  Google Brain  \\ \texttt{beenkim@google.com} } ]

\begin{abstract}
  Interpretability is an important area of research for safe deployment of machine learning systems. One particular type of interpretability method attributes model decisions to input features. Despite active development, quantitative evaluation of feature attribution methods remains difficult due to the lack of ground truth: we do not know which input features are in fact important to a model. In this work, we propose a framework for Benchmarking Attribution Methods (BAM) with a priori knowledge of \textit{relative} feature importance. BAM includes 1) a carefully crafted dataset and models trained with known relative feature importance and 2) three complementary metrics to quantitatively evaluate attribution methods by comparing feature attributions between pairs of models and pairs of inputs. Our evaluation on several widely-used attribution methods suggests that certain methods are more likely to produce false positive explanations---features that are incorrectly attributed as more important to model prediction. We open source our dataset, models, and metrics.
\end{abstract}

\section{Introduction}
The output of machine learning interpretability methods is often assessed by humans to see if one can understand it. While there is much value in this exercise, qualitative assessment alone can be vulnerable to bias and subjectivity as pointed out in~\citet{Adebayo18, Kim18}. Just because an explanation makes sense to humans does not mean that it is correct. For instance, edges of an object in an image could be visually appealing but have nothing to do with prediction. Assessment metrics of interpretability methods should capture any mismatch between interpretation and a model's rationale behind prediction.

\citet{Adebayo18} explores when certain saliency maps (a type of attribution-based explanation) fail to reflect true feature importance. Specifically, they found that certain attribution methods present visually identical explanations when a trained model's parameters are randomized (and hence features are made irrelevant to prediction). This suggests that attribution methods are making mistakes, perhaps by assigning high attributions to unimportant features (false positives) and low attributions to important features (false negatives). In this work, we focus on studying the false positive set of mistakes.

\begin{figure*}[ht]
  \centering
  \includegraphics[width=1.\linewidth]{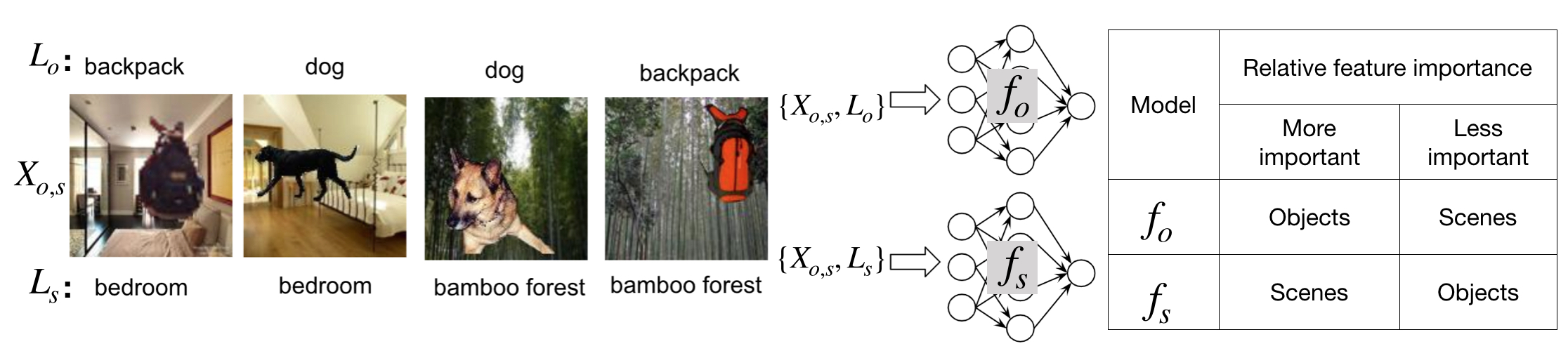}
  \caption{\textbf{BAM dataset examples and BAM models.} The object neural network ($f_o$) is trained with object labels ($L_o$) and the scene neural network ($f_s$) is trained with scene labels ($L_s$).}
  \label{fig:dataset_demo}
\end{figure*}

One way to test false positives is to identify unimportant features and expect their attributions to be zero. In reality, we do not know the \textit{absolute} feature importance (exactly how important each feature is). We can, however, control the \textit{relative} feature importance (how important a feature is to a model relative to another model) by changing the frequency certain features occur in the dataset. In this work, we propose a framework for Benchmarking Attribution Methods (BAM), which includes 1) a semi-natural dataset and models trained with known relative feature importance (Figure~\ref{fig:dataset_demo}) and 2) three metrics for quantitatively evaluating attribution methods. Given relative feature importance from our dataset, our metrics compare attributions between pairs of models (\textit{model dependence}) and pairs of inputs (\textit{input dependence} and \textit{input independence}). Our evaluation of six feature-based and one concept-based attribution methods suggests that certain methods are more likely to produce false positive explanations by assigning higher attributions to less important features. We additionally show that the rankings of attribution methods differ under different metrics; hence, whether a method is good depends on the final task and its metric of choice.

\section{Related work}
Prior work on evaluating attribution methods generally falls into four categories: 1) measuring (in)sensitivity of explanations to model and input perturbations, 2) inferring correctness of explanation from accuracy drop upon feature removal, 3) evaluating explanations in a controlled setting with some knowledge of feature importance, and 4) evaluating explanations based on human visual assessment. Our work shares characteristics with the first three: with the knowledge of relative feature importance, we measure when and how attributions should or should not change given models or inputs changing in a controlled way. Different from 4), we only focus on evaluating correctness of attributions, not on how humans understand explanations; our correctness test is a pre-check to more expensive human-in-the-loop evaluations.

\paragraph{Evaluating sensitivity of explanations.} This set of work measures how much interpretations change when models~\citep{Heo19, Adebayo18} or inputs~\citep{Ghorbani18, Alvarez18} change. \citet{Adebayo18} randomized model parameters so that features relevant to the original model are irrelevant to the randomized model. We instead train two models with different labels so that some features are more important to one model than the other. One of our metrics also measures a notion of robustness in explanations (i.e., when attributions should stay invariant) similar to \citet{Alvarez18} and \citet{Ghorbani18}. Our input perturbation, however, is semantically meaningful (e.g., adding a dog to an image). Semantically meaningful perturbations are well-suited for testing attribution methods because they can mislead humans if attributed incorrectly.

\paragraph{Evaluating correctness of explanations.} \citet{Samek17} and its subsequent variations~\citep{Fong17, Ancona18, Hooker18} remove highly attributed input features and measure the resulting performance degradation. Without retraining the model, the modified inputs may fall outside of the training data manifold. As a result, it is hard to decouple whether the accuracy drop is due to out-of-distribution data or due to good feature attributions. Retraining, on the other hand, results in a model different from the original one being explained. We enable method evaluation on the original model using in-distribution inputs.

\paragraph{Evaluating with knowledge of feature importance.}
Most relevant to our work is the controlled experiment in~\citet{Kim18}, where the authors created a simple dataset and trained models such that important pixels are known. While~\citet{Kim18} compared their method to saliency maps, no quantitative metrics were given; their results were qualitatively evaluated by human subjects. Our work develops quantitative metrics in a finer-grained setting with semi-natural images.

\paragraph{Evaluating with humans in the loop.} This approach either asks humans to use interpretability methods to perform a task or evaluates how often humans can correctly predict model behavior after seeing the explanations~\citep{Lakkaraju16, Ribeiro16, Doshi17, Lage18, Narayanan18, Poursabzi18}. Our work is not meant to replace but to complement human-in-the-loop evaluations. We provide an efficient pre-check to make sure that methods correctly assign lower attributions to less important features before involving humans in the evaluation loop.

\begin{figure*}[ht]
\centering
  \centering
  \includegraphics[width=.9\linewidth]{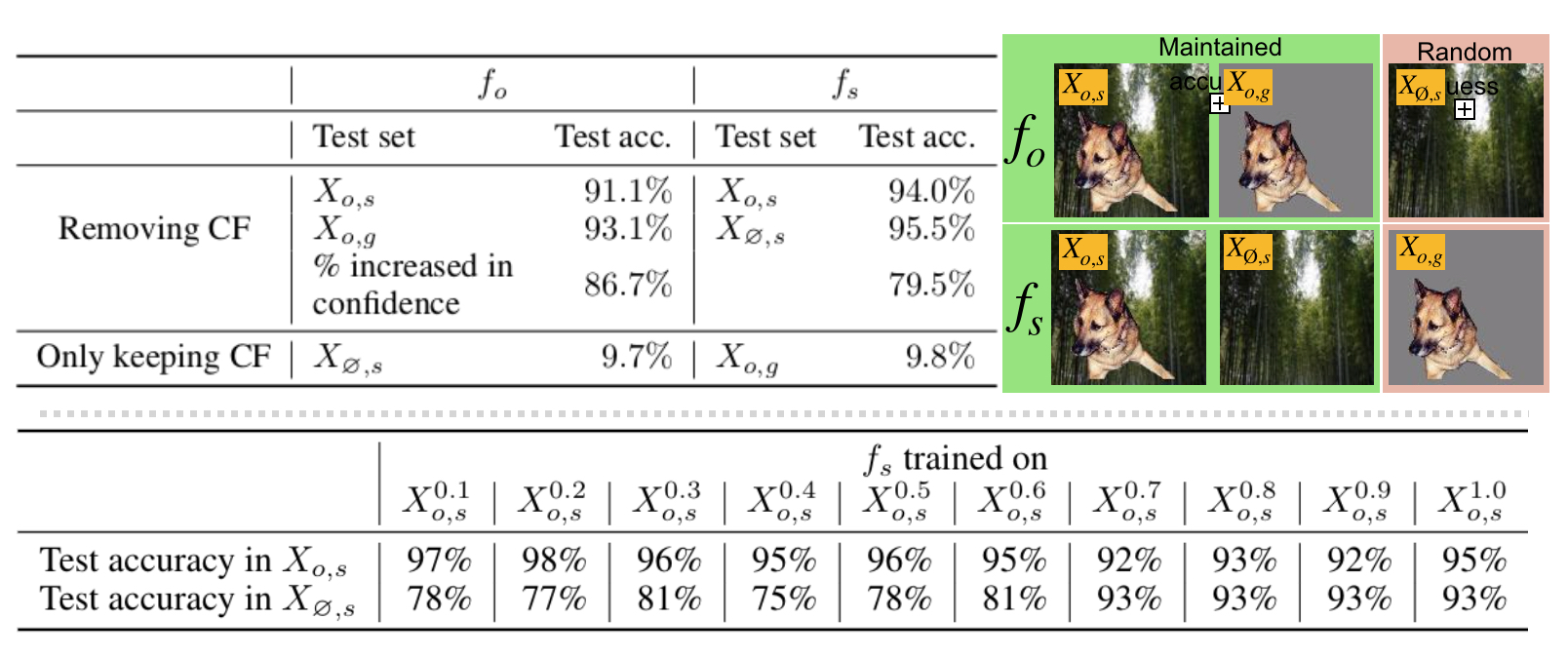}
\caption{[Top] Verifying relative feature importance between $f_o$ and $f_s$. Objects are more important to $f_o$ than they are to $f_s$. Scenes are more important to $f_s$ than to $f_o$. [Bottom] Test accuracy of bamboo forest with and without dog \CF~on models trained with $\{X_{o,s}^k\}$ for $k \in \{0.1, \dots, 1.0\}$.}
  \label{fig:dataset_gray_table:dataset_acc} 
\end{figure*}

\section{BAM dataset and models}
\label{sec:dataset}
Our goal is to evaluate correctness of feature attributions with respect to their relative importance. For instance, if we paste a gray square to every training image for all classes, we expect this square to matter less to model predictions than the original image region being covered. The same expectation should hold if we instead paste a dog to every image (given that the original images do not contain anything similar to dogs). As a result, any explanations that assign higher attributions to dogs than to original pixels covered by dogs are false positives. False positive attributions of dogs have a higher chance of misleading humans because dogs are semantically meaningful. 

Based on this idea, we propose the BAM framework which includes 1) the BAM dataset and BAM models trained with known relative feature importance and 2) BAM metrics that quantitatively evaluate attribution methods (with and without BAM models). In this section, we describe the dataset and models, which are open sourced at \url{https://github.com/anonymous}.

\subsection{BAM dataset construction}
We construct the BAM dataset by pasting object pixels from MSCOCO~\citep{Lin14} into scene images from MiniPlaces~\citep{Zhou17}, as shown in Figure~\ref{fig:dataset_demo}. An object is rescaled to between \nicefrac{1}{3} to \nicefrac{1}{2} of a scene image and is pasted at a randomly chosen location. Each resulting image has an object label ($L_o$) and a scene label ($L_s$). Either can be used to train a classifier. BAM dataset has $10$ object classes (e.g., backpack, bird, and dog) and $10$ scene classes (e.g., bamboo forest, bedroom, and corn field). Every object class appears in every scene class and vice versa. There is a total of $100k$ images ($1k$ images per object-scene class pair). Scene images that originally contain objects from BAM's object categories are filtered (e.g., no scenes originally have dogs). We denote this set as $X_{o,s}$.

\subsection{Common feature (\CF) and commonality}
We define \commonF~(\CF) as a set of pixels with some semantic meaning (e.g., looks like a dog) that commonly appear in all examples of one or more classes. The ratio between the number of classes a \CF~appears in and the total number of classes is the commonality ($k$) of that \CF. For example, a dog \CF~where dog pixels appear in all images of bamboo forest ($k=0.1$) is less common than another dog \CF~where dog pixels appear in all images of bamboo forest, bedroom, and corn field ($k=0.3$). We denote $X_{o,s}^k$ as a set of inputs where $k$ percent scene classes have a particular \CF. We create $10$ such sets, $\{X_{o, s}^k\}$ for $k \in \{0.1, ..., 1.0\}$, using dog pixels as \CF. Each set has $10k$ images with scene labels.

\subsection{Relative importance of BAM models}
\label{sec:relative_feature_importance_models}
We consider two scenarios where feature importance between models differ: 1) two classifiers trained on different labels ($L_o$ and $L_s$), and 2) a set of scene classifiers trained with dog \CF~of different $k$ values.

\paragraph{Training two classifiers with coarse-grained \CF.}
We first train two classifiers using $X_{o,s}$. $f_s$ is the classifier trained with scene labels $L_s$, and $f_o$ is the classifier trained with object labels $L_o$. As a result, objects are considered coarse-grained \CF~(with $k=1.0$) to $f_s$. Intuitively, objects should be significantly more important to $f_o$ where they are targets of classification than to $f_s$ where they are \CF. To verify this intuition, we remove object pixels from $X_{o,s}$ while leaving the original scenes intact (denoted as $X_{\varnothing,s}$). The accuracy of $f_o$ drops to a random guess, but the accuracy of $f_s$ does not drop at all, as shown in Figure~\ref{fig:dataset_gray_table:dataset_acc}. Similarly, when we remove scenes from $X_{o,s}$ by filling background with gray pixels (denoted as ${X_{o,g}}$), the accuracy of $f_s$ drops to a random guess but the accuracy of $f_o$ does not drop. Knowing that objects are significantly more important to $f_o$ than to $f_s$, we can test whether methods assign noticeably higher attributions to objects in $f_o$ than in $f_s$.

\paragraph{Training scene classifiers with fine-grained \CF s.}
We enable a finer-grained control over relative feature importance between models by training classifiers on sets of inputs whose \CF s differ in commonality $k$. In particular, we train $10$ scene classifiers on $\{X_{o,s}^{k}\}$ for $k \in \{0.1, \dots, 1.0\}$ with dog \CF. By changing $k$, we can control dog \CF 's relative importance to the model. Intuitively, dog \CF~with a lower $k$ is more important to prediction than dog \CF~with a higher $k$. We verify this intuition by removing dog \CF~and measuring the accuracy drop of bamboo forest (the randomly chosen class that always has dog \CF~during training for different $k$). As shown in Figure~\ref{fig:dataset_gray_table:dataset_acc} (bottom), the accuracy drop generally decreases as $k$ increases. This indicates that the relative importance of dog \CF~also decreases as $k$ increases. With dog \CF 's relative importance, we can evaluate methods by comparing \CF~attributions across the set of models: a method should assign a higher attribution to the \CF~that is less common.

\subsection{Relative importance of BAM inputs}
\label{sec:relative_feature_importance_inputs}
After training $f_s$ on $X_{o,s}$ using $L_s$, we observe in Figure~\ref{fig:dataset_gray_table:dataset_acc} that the test accuracy of $f_s$ is higher on $X_{\varnothing, s}$ than on $X_{o,s}$ (higher accuracy when object \CF~is removed). This implies that object \CF~is less important to $f_s$ than the original scene pixels they cover. In addition, we measure the difference in classification confidence when object \CF~is removed from the input. As shown in Figure~\ref{fig:dataset_gray_table:dataset_acc}, the majority ($86.7\%$) of the correctly classified inputs have increased classification confidence (in the rest classification confidence roughly stays the same). With relative importance between inputs with and without \CF, we can test if methods assign lower attributions to object \CF~than the scene pixels being covered.

 So far we have considered when feature importance between inputs are different. We can create another scenario (without requiring the BAM dataset or BAM models) where feature importance between inputs are similar. This allows us to test if methods assign similar attributions to inputs whose features have similar importance.
 
\section{Metrics with/without BAM dataset} \label{sec:metric}
We propose three complementary metrics to evaluate attribution methods: model contrast score (\MCS), input dependence rate (\IDR), and input independence rate (\IIR). These metrics measure the difference in attributions between two models given the same input (\MCS), between two different inputs to the same model (\IDR), and between two functionally similar inputs (\IIR). The measurements can then be compared against the relative feature importance between models and inputs established in Section~\ref{sec:dataset}. \MCS~and \IDR~require the BAM dataset and BAM models, whereas \IIR~applies to \textit{any} model given \textit{any} inputs. Our metrics are by no means exhaustive---they only focus on false positive explanations, but they demonstrate how differences in attributions can be used for method evaluation, enabling development of additional metrics. BAM metrics are cheap to evaluate and therefore can serve as a pre-check before conducting expensive human-in-the-loop evaluations.

\subsection{Setup}
First, we define a way to compute the average attribution that a saliency method assigns to an image region. Taking the average is acceptable because we only care about \textit{relative} feature importance. We denote $e \in \mathbb{R}^d$ as a saliency map whose values are normalized to $[0, 1]$, and $I_c \in \mathbb{R}^d$ as a binary mask where pixels inside a region $c$ have a value of $1$, and $0$ everywhere else. Given an input image $x \in \mathbb{R}^d$ and a model $f: \mathbb{R}^d \mapsto \mathbb{R}^m$, we define $g_c  \in \mathbb{R}$ as the average attribution of pixels in region $c$, namely
$$
  g_c(f, x) = \frac{1}{\sum I_c}\sum{e(f, x) \odot I_c}.
$$
We further define concept attribution, $G_c \in \mathbb{R}$, as the average of $g_c$ over a set of correctly classified inputs ($X_{\textit{corr}}$), where region $c$ represents some human-friendly concept (e.g., objects) in each input:
$$
  G_c(f, X) = \frac{1}{|X_{\textit{corr}}|}\sum\limits_{x \in X_{\textit{corr}}}{g_c(f, x)}.
$$
We only consider correctly classified inputs so that classifiers' mistakes are not propagated to attribution methods. Note that the above definition of $G_c$ is only needed for feature-based explanations but not for concept-based explanations that compute their own $G_c$ such as TCAV~\citep{Kim18}. Our metrics also apply to these methods as long as they provide a $G_c$ value between $[0, 1]$. Now we define our three metrics.

\subsection{Model contrast score (\MCS)} \label{sec:model_dependence}
Once we have a \CF~that appears in every class, one option is to directly compare $G_c$ values of that \CF~across methods, and expect $G_c$ to be small. However, there is a catch: a meaningless $e$ that always assigns 0 (unimportant) to all features would seem to perform well. Even without such a bogus $e$, two attribution methods may operate on two different scales, so directly comparing $G_c$ could be unfair. 

Section~\ref{sec:relative_feature_importance_models} introduced relative feature importance between models---some input features are more important to one model than to another model. For example, dog pixels more important to $f_o$ than to $f_s$. We define model contrast score (\MCS) as the difference in concept attributions between a model that considers a concept more important ($f_1$) than another model ($f_2$):
$$
\textrm{\MCS} = G_c(f_1, X_{\textit{corr}}) - G_c(f_2, X_{\textit{corr}}).
$$

We expect \MCS~to be the largest when $f_1 = f_o$, $f_2 = f_s$, $X_{\textit{corr}} \subset X_{o,s}$, and $c$ corresponds to objects in the BAM dataset. This measures how differently objects are attributed between $f_o$ and $f_s$ --- between when they are targets of classification (more important) and when they are \CF~(hence less important). We also compute a series of \MCS~scores by training $f_1$ using one of $\{X_{o,s}^k\}$ for $k \in \{0.1, \dots, 0.9\}$, and $f_2$ using $X_{o,s}^{1.0}$. This gives us a spectrum of contrast scores where the dog \CF~is important to a different degree.

\subsection{Input dependence rate (\IDR)}
While \MCS~compares attributions between models, one may also be interested in how a method performs on a single model. Section~\ref{sec:relative_feature_importance_inputs} introduced relative feature importance between inputs---a \CF~(with $k=1.0$) is less important to a model (trained with that \CF) than regions covered by that \CF. Based on this knowledge, we can compare attributions of two inputs with and without~\CF, and expect $g_c$ of the \CF~region to be smaller when \CF~is present. We define input dependence rate (\IDR) as the percentage of correctly classified inputs where \CF~is attributed less than the original regions covered by \CF. Formally,
$$ \textrm{\IDR} = \frac{1}{|X_{cf}|}\qquad \quad \; \sum\limits_{\mathclap{(x_{cf}, x_{\neg cf}) \in (X_{cf}, X_{\neg cf})}}{\mathds{1}(g_{c}(f, x_{cf}) < g_{c}(f, x_{\neg cf})}),$$

where $x_{cf}$ and $x_{\neg cf}$ are inputs with and without \CF. Since we only require $c$ to be an input region, we can still compute $g_c$ of that region even if \CF~is absent. In the BAM framework, we have $X_{cf} = X_{o, s}$, $X_{\neg cf} = X_{\varnothing, s}$, and $f = f_s$. A high \IDR~means that an attribution method is more likely to correctly attribute \CF. Intuitively, $1-\textrm{\IDR}$ is the false positive rate: out of 100 inputs, how many of their explanations assign higher attributions to less important features, thereby misleading human interpretation. 

\subsection{Input independence rate (\IIR)} \label{sec:method_ii}
While \IDR~considers two inputs that should be attributed differently, input independence rate (\IIR) expects similar attributions between two inputs that lead to the same model output (logit scores for all classes). Given any model $f$ and any input $x$, we propose an optimization-based approach to compute a set of connected pixels $\delta$, such that $f(x) \approx f(x+\delta)$. Because $x$ and $x+\delta$ have similar outputs, we call $x$ and $x+\delta$ functionally similar. This $\delta$, in fact, can have a large norm due to the excessive invariance of neural networks found in~\citet{Jacobsen18, Engstrom19}. Specifically, we use gradient descent to optimize the following objective:
$$ \arg\min_\delta \|f(x+\delta)-f(x)\|_2 - \eta_1\|\delta\|_2 + \pazocal{R},$$
where $-\eta_1\|\delta\|_2$ avoids the trivial solution of $\delta=0$. $\pazocal{R}$ is a regularization term that encourages $\delta$ to look natural (see details in Appendix). This optimization is applicable to any models where gradients are accessible. When $\delta$ aligns well with a human-friendly concept, a false positive explanation that highly attributes $\delta$ could be dangerously misleading, as the output is the same with or without $\delta$. Hence, we initialize $\delta$ from pixels of a dog, and only modify those pixels within a small $L_2$ distance of initialization. Figure~\ref{fig:ii_demo} shows an example post-optimization with $\delta$ still looking like a dog. For this particular example, we have $\|f(x+\delta)-f(x)\|_2 = 0.2$, which is magnitudes smaller than $\|f(x)\|_2 = 31.9$. We verify that such a $\delta$ exists for any given input.

Next, we compute $g_c$ for pairs of $x$ and $x + \delta$, where $x$ is drawn from the original scene images ($X_{\varnothing,s}$). When $\delta$ is absent, $g_c$ is computed over the region where $\delta$ would have been. We expect $g_c$ to only change within a visually imperceptible threshold $t$, above which humans would notice the difference in attribution when $\delta$ is added. We define \IIR~as the percentage of images where the difference in $g_c$ with and without $\delta$ is less than $t$:
$$
\textrm{\IIR} = \frac{1}{|X_{\textit{corr}}|}\sum\limits_{x \in X_{\textit{corr}}} \mathds{1}(\frac{|g_c(f, x + \delta) - g_c(f, x)|}{g_c(f, x)} < t). $$
Here $f$ is trained on $X_{\varnothing,s}$, but it can be any model. Intuitively, $1-\textrm{\IIR}$ is the false positive rate: out of 100 inputs, how many of their explanations would draw attention to $\delta$ which has no functionality, thereby misleading human interpretation. We note that $t$ depends on human subjects and is application specific.

\begin{figure*}[ht]
\centering
\begin{minipage}{.48\textwidth}
  \centering
  \includegraphics[width=1.\linewidth]{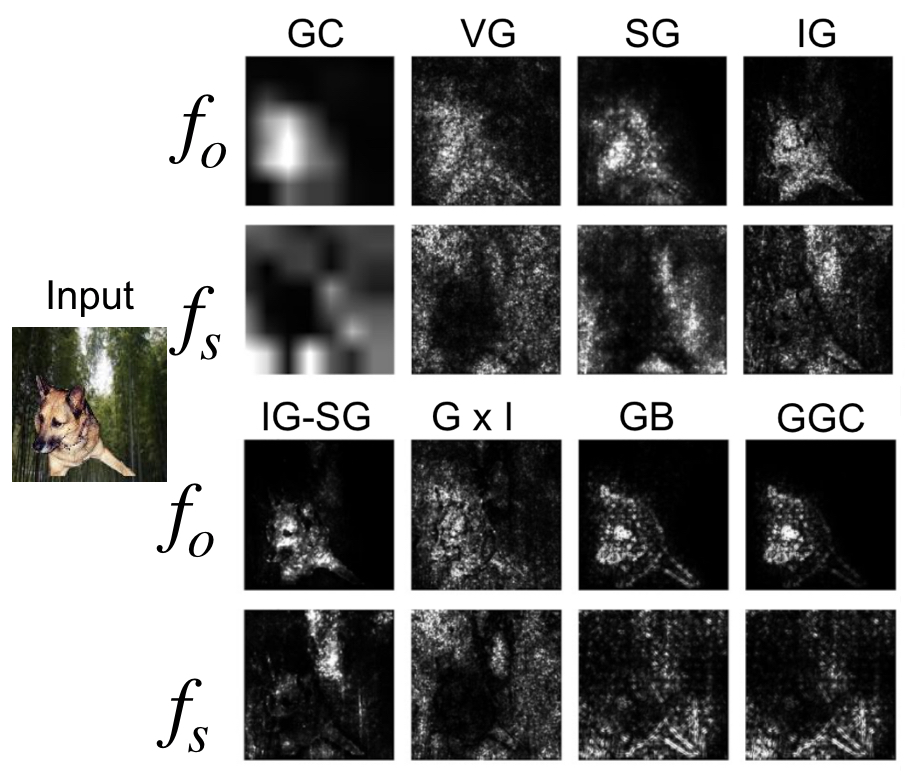}
  \caption{
  An example of saliency map visualizations for $f_o$ and $f_s$. From qualitative examination alone, it is hard to rank method performance.
  }
  \label{fig:mc_demo}
\end{minipage}
  \hfill
\begin{minipage}{.48\textwidth}
  \centering
  \includegraphics[width=1.\linewidth]{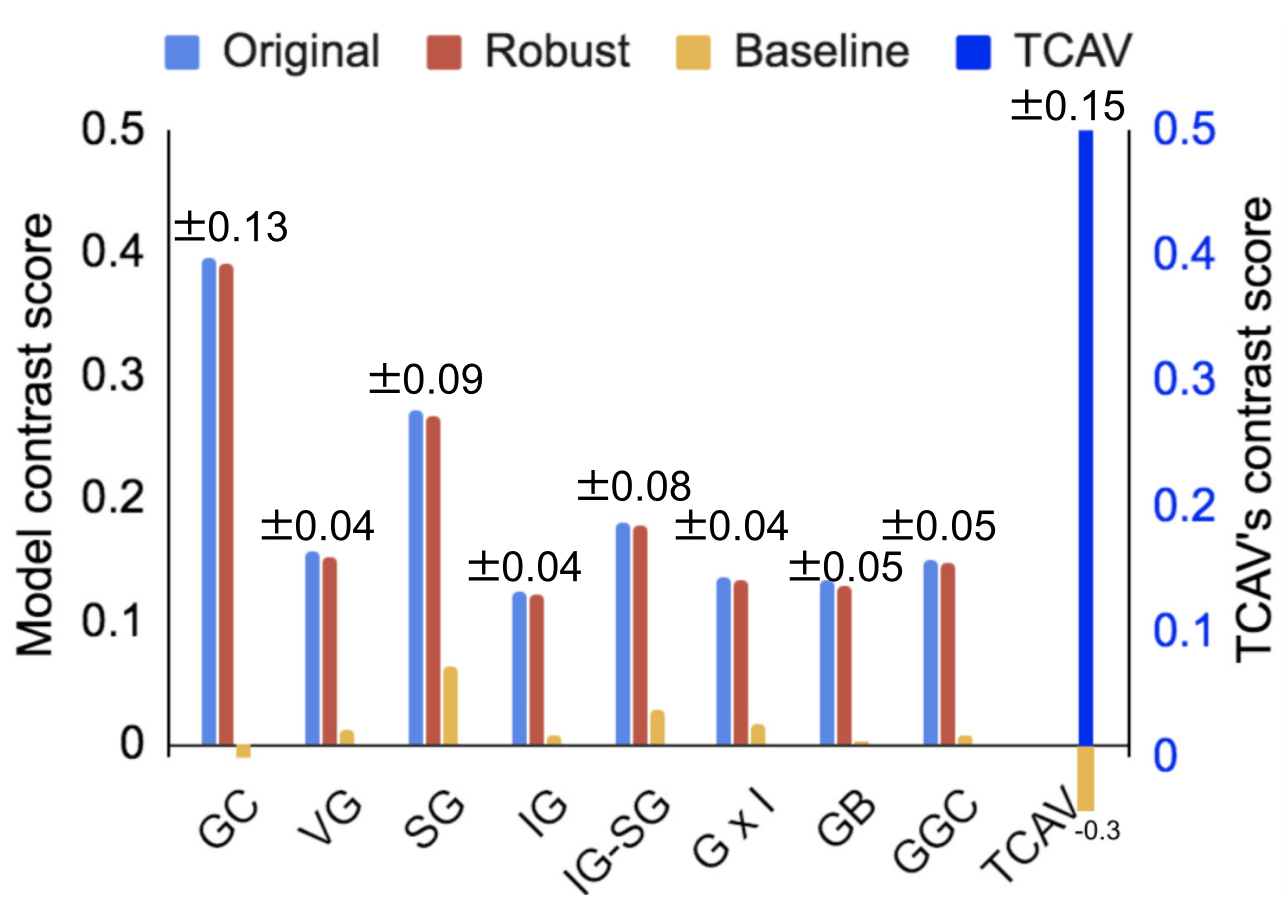}
  \caption{\MCS~between $f_o$ and $f_s$. Blue bars are measurements from the original BAM dataset. Red bars show robustness of this measure. Yellow bars are baselines. Numbers on top are standard deviations. Higher \MCS~is better.}
  \label{fig:mc_bar}
\end{minipage}
\end{figure*}

\section{Evaluation with/without BAM}
With the BAM dataset and BAM models from Section~\ref{sec:dataset} and BAM metrics from Section~\ref{sec:metric}, we evaluate feature-based attribution methods including GradCAM (GC)~\citep{Selvaraju16}, Vanilla Gradient (VG)~\citep{Simonyan13,Erhan09,Baehrens10}, SmoothGrad (SG)~\citep{Smilkov17}, Integrated Gradient (IG)~\citep{Sundararajan17}, and Guided Backpropagation (GB)~\citep{Springenberg14}. We also consider Gradient x Input (GxI), as many methods use GxI to visualize attributions. Our saliency map visualization follows the same procedure as~\citet{Smilkov17}, except that we only use positive pixel attributions, as our work focuses on false \textit{positives}. We also use \MCS~to evaluate a concept-based attribution method (TCAV).

In summary, our results suggest that 1) certain methods (such as GB) tend to assign similar attributions despite changes in underlying feature importance, 2) GC and VG are the least likely to produce false positive explanations, and 3) our metrics are indeed complementary---VG, for example, has high \IDR~and \IIR~but low MCS. Therefore, applications should choose attribution methods based on the metrics they care about the most (e.g., higher contrast between explanations, fewer false positive explanations, or robust explanations in the presence of functionless changes in the input).

\subsection{Attributions between models with MCS}
\MCS~offers two sets of evaluations---between two models trained using different labels and between a set of models trained with \CF s of different commonality $k$. The rankings from the two sets of evaluations are similar: TCAV and GC have the best MCS, outperforming all the other methods.

\paragraph{\MCS~with coarse-grained \CF.} One image from the BAM dataset and its saliency maps computed on $f_o$ and $f_s$ are shown in Figure~\ref{fig:mc_demo}. With qualitative examination alone, it is hard to see which method is better than others. We compute \MCS~between $f_o$ and $f_s$ over $10$k images from $X_{o,s}$. The quantitative results are shown in Figure~\ref{fig:mc_bar}. TCAV has the highest \MCS, immediately followed by GC. By applying smoothing to VG, SG improves \MCS, outperforming the rest methods. We confirm that \MCS~is robust to the scale and location of objects (red bars). The baseline (yellow bars) is calculated using random $I_c$. This means how much attributions of randomly chosen features differ between $f_o$ and $f_s$ (see details in Appendix). \MCS~for TCAV is computed by taking the difference in object TCAV scores between $f_o$ and $f_s$.

\begin{figure*}[ht]
\centering
\begin{minipage}{.48\textwidth}
  \centering
  \includegraphics[width=1.\linewidth]{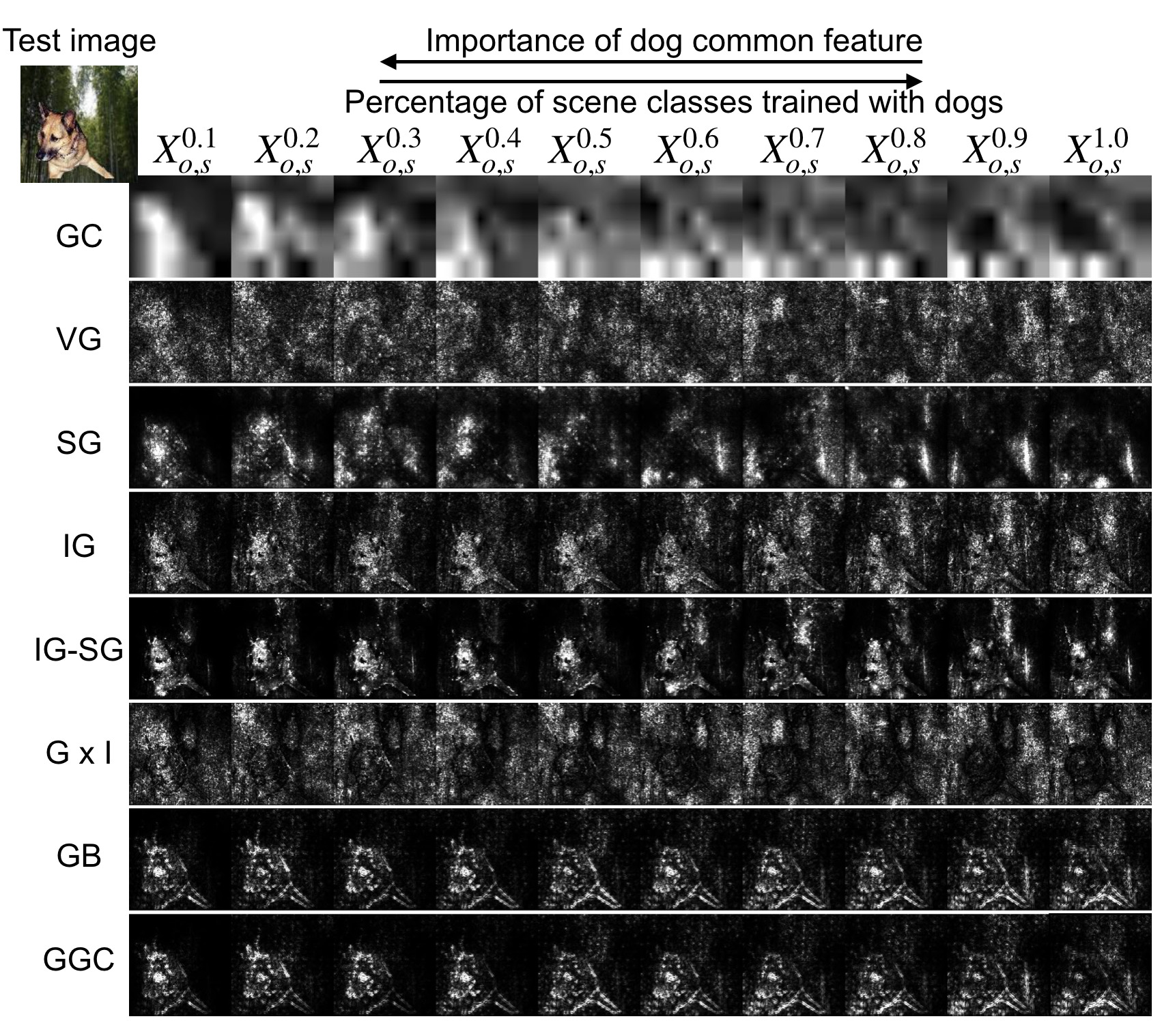}
  \caption{An example of saliency map visualizations for models trained with \CF s of different $k$. $k$ increases from left to right. A larger contrast among each row is better. (Full size figure in Appendix.) }
  \label{fig:rmc_demo}
\end{minipage}
  \hfill
\begin{minipage}{.48\textwidth}
  \centering
  \includegraphics[width=1.\linewidth]{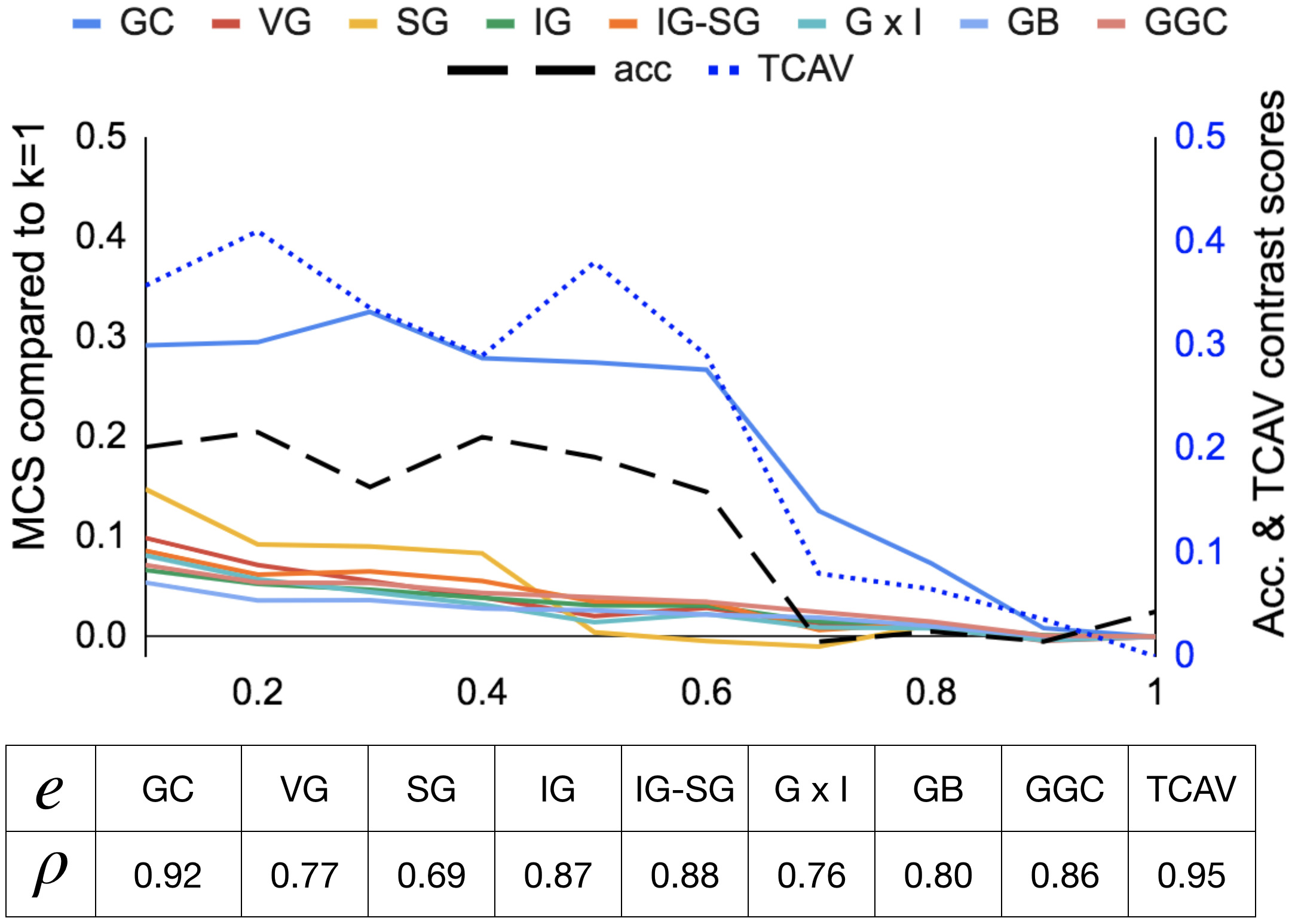}
  \caption{[Top] \MCS~between $\{X_{o,s}^k\}$ for $k \in \{0.1, \dots, 1.0\}$ and $X_{o,s}^{1.0}$ as $k$ increases. The dashed black line is the accuracy drop when \CF~is removed during testing. The dotted blue line is the series of \MCS~for TCAV. [Bottom] Pearson correlation coefficients ($\rho$) between each method's \MCS~and the accuracy drop. A higher correlation is better.}
  \label{fig:rmc_bar}
\end{minipage}
\end{figure*}

\paragraph{\MCS~with fine-grained \CF s} We also compute a series of \MCS~between a classifier trained on $X_{o,s}^{1.0}$ and a set of classifiers trained on $\{X_{o,s}^k\}$ for $k \in \{0.1, \dots, 1.0\}$, as described in Section~\ref{sec:relative_feature_importance_models}. As the commonality $k$ of dog \CF~increases from left to right in Figure~\ref{fig:rmc_bar} (top), TCAV and GC closely follow the trend of accuracy drop upon \CF~removal (dashed black line). Visual assessment of GC in Figure~\ref{fig:rmc_demo} supports this observation: the first row shows that GC's dog attribution decreases as $k$ increases. All methods except TCAV and GC change at a much smaller scale. In particular, GB evolves minimally with the edges of the dog always being visible (second to last row of Figure~\ref{fig:rmc_demo}), similar to the findings in~\citet{Adebayo18,Nie18}. Figure~\ref{fig:rmc_bar} (bottom) additionally shows the Pearson correlation coefficients between each method's \MCS~and the accuracy drop upon \CF~removal. TCAV achieves the highest correlation closely followed by GC. 

\subsection{Attributions between inputs with IDR}
\begin{figure*}[ht]
\centering
\begin{minipage}{.48\textwidth}
  \centering
  \includegraphics[width=1.\linewidth]{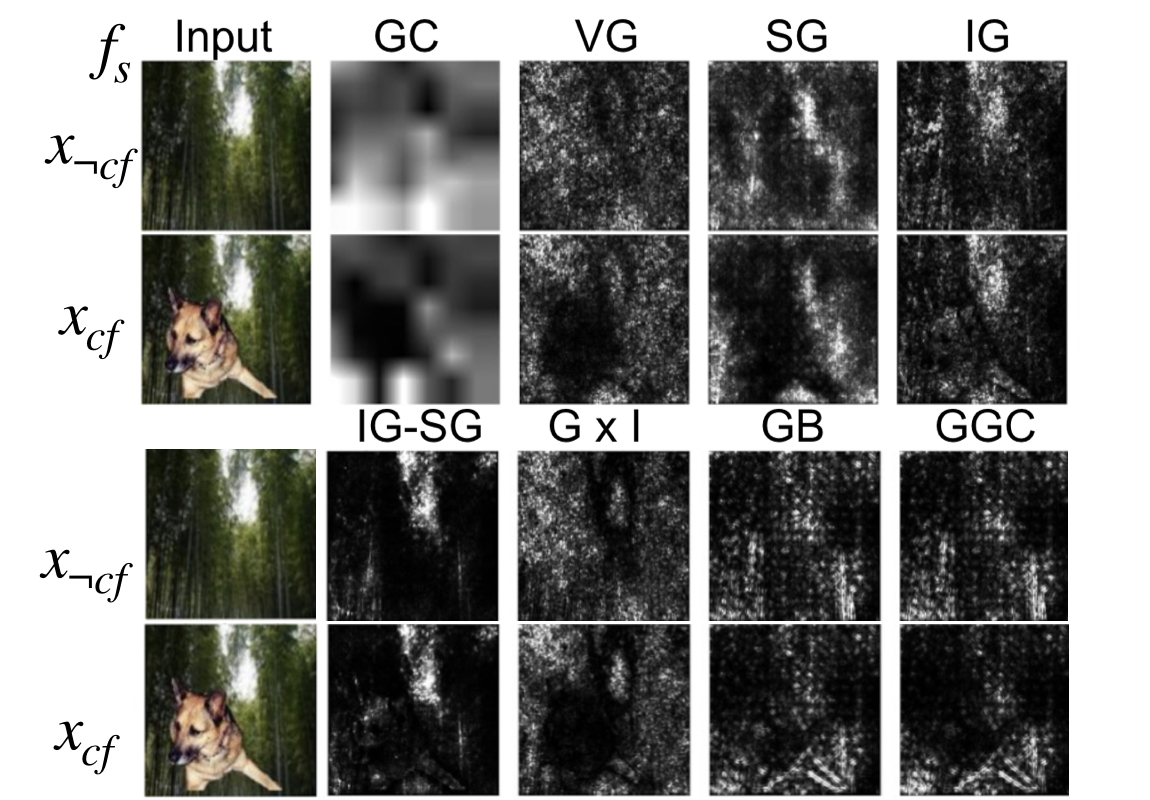}
  \caption{Saliency maps with and without \CF~on $f_s$. The dog \CF~is less important than the scene region being replaced.}
  \label{fig:id_demo}
\end{minipage}
  \hfill
\begin{minipage}{.48\textwidth}
  \centering
  \includegraphics[width=1.\linewidth]{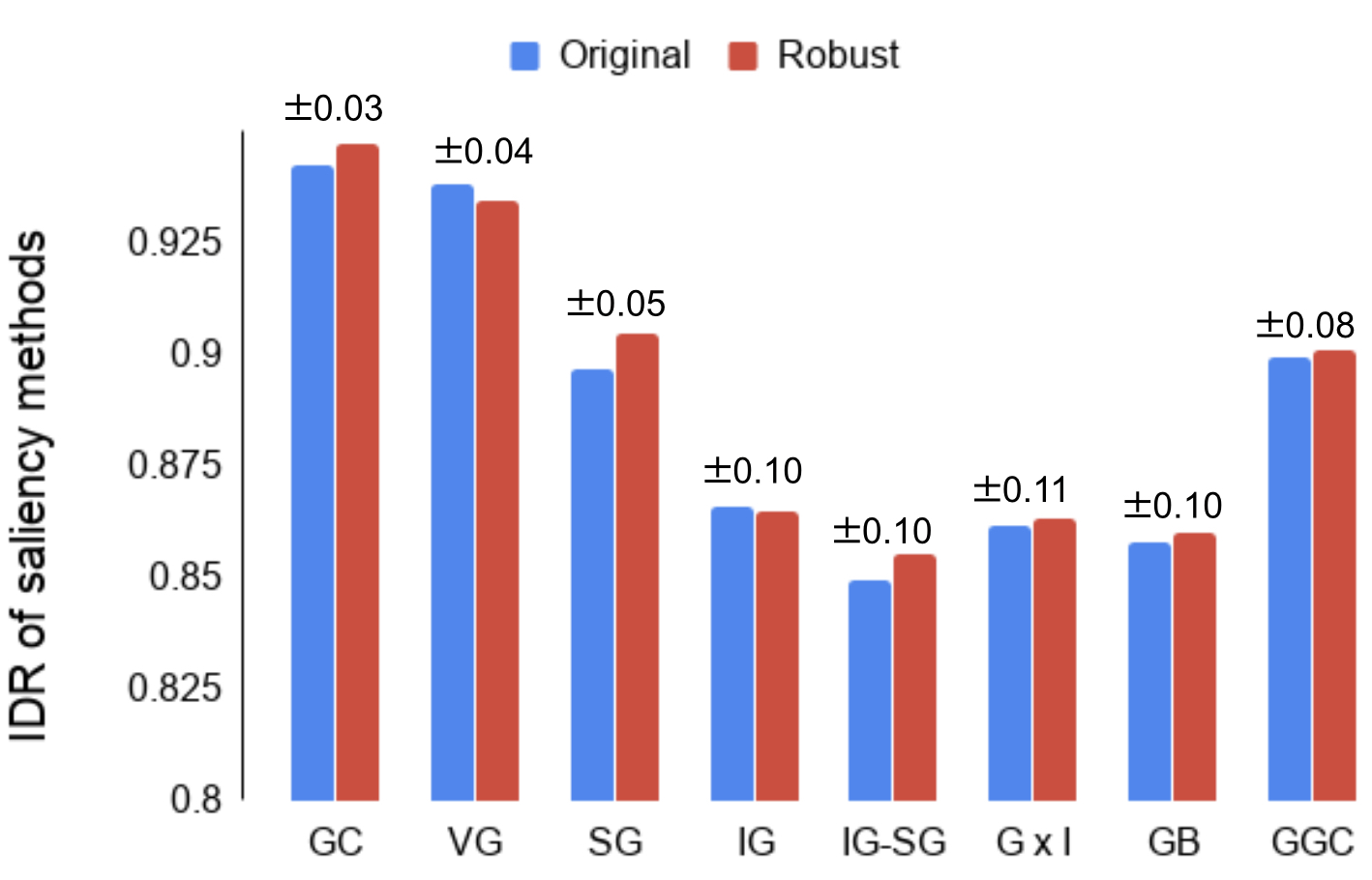}
  \caption{\IDR~results for $(X_{cf}, X_{\neg cf})$. Higher \IDR~is better. Red bars are robustness test of the metric. Numbers on top are standard deviations over 10 trials of 100 image pairs each. Baseline is 50\%.}
  \label{fig:id_bar}
\end{minipage}
\end{figure*}

Now we assess attribution difference between inputs to $f_s$ with and without \CF. From visualization alone (Figure~\ref{fig:id_demo}), it is again hard to rank the saliency methods. The quantitative measurement of \IDR~in Figure~\ref{fig:id_bar} shows that GC and VG have the most correctly attributed \CF---the least amount of false positive explanations. We note that VG is the original gradient method; many other methods require additional steps after computing VG. This means that the cheapest method offers nearly the best performance (also suggested by~\citet{Adebayo18}). The baseline of \IDR~is around 50\% (measured using random $I_c$). In applications where low false positive rate is critical, GC and VG are better choices than other methods. \IDR~is not applicable to TCAV, as TCAV is a global method.

\begin{figure*}[h!]
\centering
\begin{minipage}{.48\textwidth}
  \centering
  \includegraphics[width=1.\linewidth]{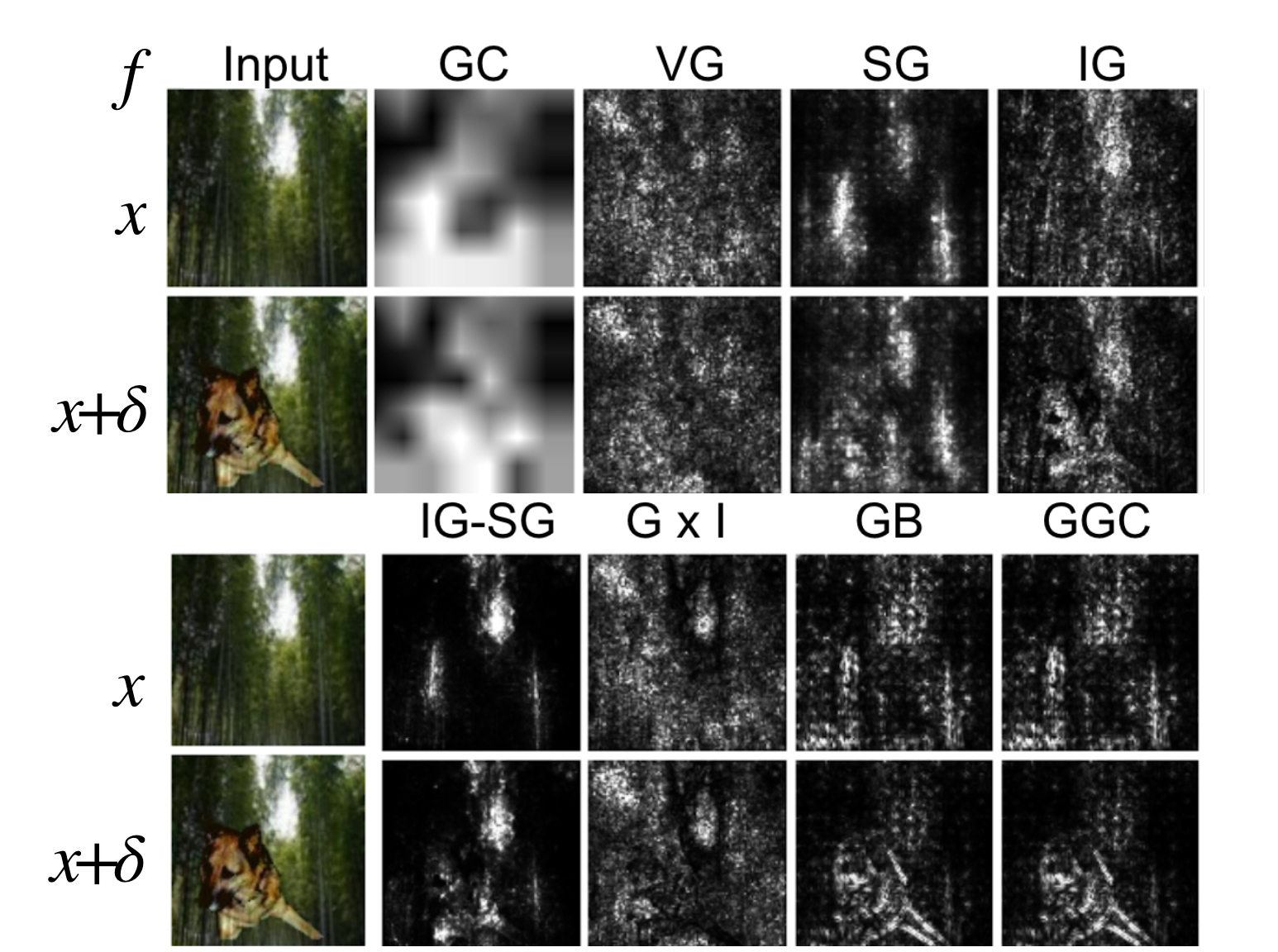}
  \caption{Saliency maps for $x$ and $x+\delta$ on any $f$. $x$ and $x+\delta$ are functionally similar.}
  \label{fig:ii_demo}
\end{minipage}
  \hfill
\begin{minipage}{.48\textwidth}
  \centering
  \includegraphics[width=1.\linewidth]{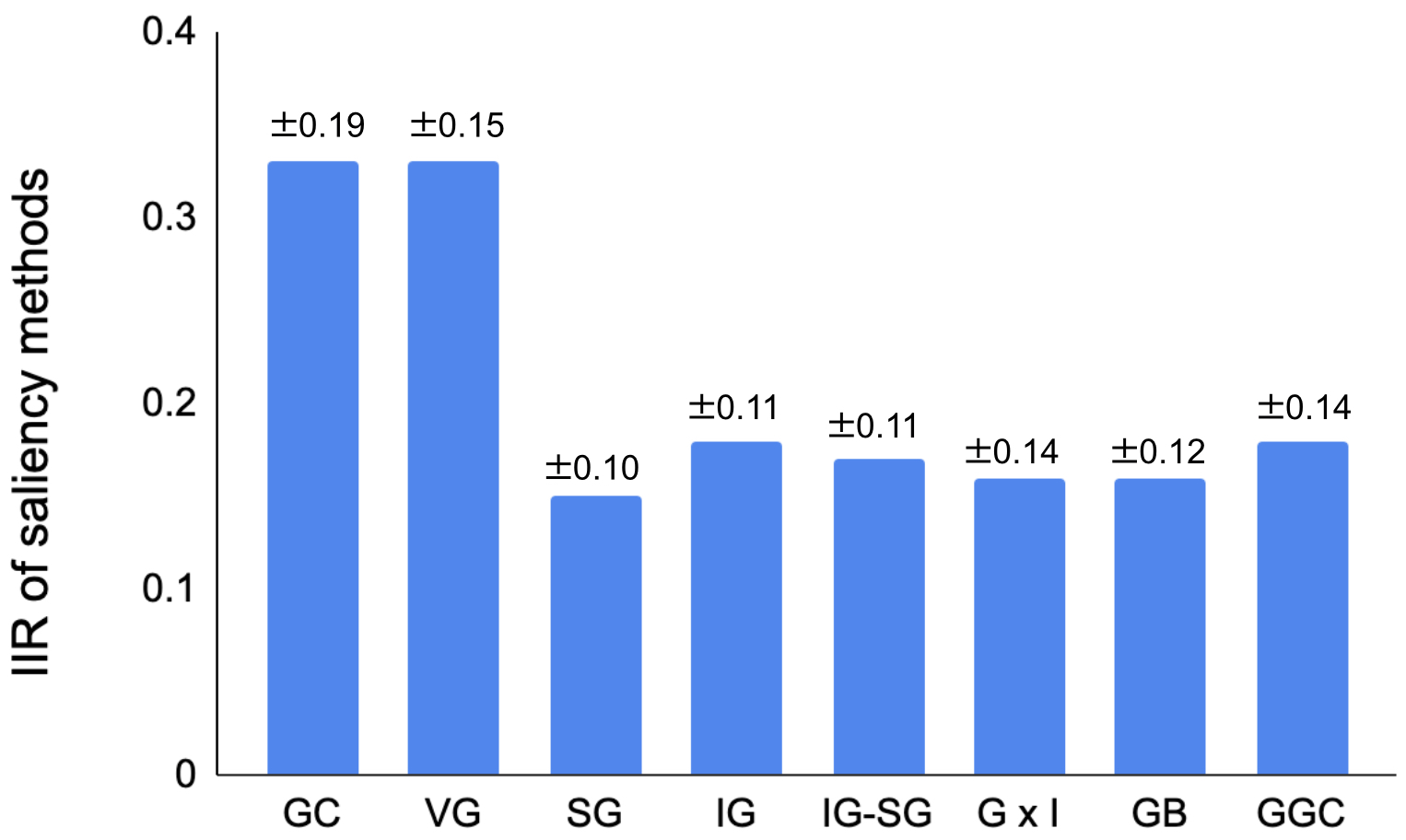}
  \caption{\IIR~results with $t=10\%$. Higher \IIR~is better. Numbers on top are standard deviations over 10 trials of 100 image pairs each.}
  \label{fig:ii_bar}
\end{minipage}
\end{figure*}

\subsection{Attributions between functionally similar inputs with IIR}
Finally, we present the \IIR~results (which could have been measured on any model given any inputs). Most of the methods, with the exception of GC and VG, incorrectly assign much higher attributions to $\delta$ for over 80\% of the examples (Figure~\ref{fig:ii_bar}). This alarming result is confirmed by visual assessments in Figure~\ref{fig:ii_demo}---the dog is clearly highlighted by IG, IG-SG, GB, and GGC. The finding that GB tends to highlight edges is consistent with \citet{Adebayo18,Nie18}. Since $\delta$ is highly visible in the input image but does not affect prediction, attributions that directly depend on the input, such as GxI and IG, are likely to be affected by $\delta$. This observation calls into question the common practice of multiplying explanations by the input image for visualization.

$t=10\%$ is used as the threshold to compute \IIR~(Section~\ref{sec:method_ii}). We visually determined that when $g_c$ changes by more than $10\%$, we could clearly see the attribution difference of the $\delta$ region in some methods. The choice of $t$ can be further calibrated by human subjects. As a reminder, \IIR~with $t=10\%$ is the percentage of inputs where adding $\delta$ does not change the attribution of the region by more than $10\%$. We use $\eta_1 = 0.01$ when computing this $\delta$, but the optimization is robust to the choice of $\eta_1$. \IIR~is not applicable to TCAV, as TCAV is a global method.

\section{Discussion}
We evaluate methods based on how well difference in attributions aligns with relative feature importance. Perturbation-based evaluations, on the other hand, remove highly attributed features and infer method performance by the resulting accuracy drop. Below we discuss the differences between our approach and perturbation tests, and how our evaluation results might generalize to real datasets.


\paragraph{Assumptions about feature importance.}Perturbation-based evaluations first run attribution methods on an input and rank features by the resulting attributions. A fixed percentage of highly ranked features are then removed, and method that leads to the highest accuracy drop is the best. Perturbation-based evaluations assume that there exist a unique set of important features whose removal would cause an accuracy drop. However, the set of important features may not be unique. The BAM framework, on the other hand, does not make assumptions about importance of individual features.



\paragraph{Computational efficiency.} Perturbation tests are computationally expensive. For instance, the cost of retraining in~\citet{Hooker18} is linear to the number of methods being examined. Tests that do not retrain the model after modifying the input such as~\citet{Fong17, Ancona18, Hooker18} still incurs a cost that is linear to the number of perturbation steps. Our test requires training a fixed number of models only once, and does not use small (and therefore costly) perturbation steps.


\paragraph{Do evaluation results on the BAM dataset generalize to real dataset?}
One might wonder how our evaluation results on the BAM dataset translates to natural images. While there is no guarantee that methods performing well on the BAM dataset will assign correct attributions to real images, BAM could be seen as a simpler and easier test to more complex scenarios. If an attribution method fails an easier test, it is also likely to fail harder tests.
Furthermore, the metrics we evaluate, such as assigning lower attribution to less important features (hence avoiding false positives), reflect good properties of attribution methods regardless of the underlying dataset.

\section{Conclusions}
There is little point in providing false explanations---evaluating explanations is as important as developing attribution methods. In this work, we take a step towards quantitative evaluation of attribution methods using relative feature importance between different models and inputs. We create and open source a semi-natural image dataset (BAM dataset), a set of models trained with known relative feature importance (BAM models), and three complementary metrics (BAM metrics) for evaluating attribution methods. Our work is only a starting point; one can also develop other measures of performance using our framework. We hope that developing ways to quantitatively evaluate attribution methods helps us choose the best metric and method for the application at hand.

\clearpage
\bibliography{references}
\bibliographystyle{main}

\clearpage
\appendix
\input{appendix}

\end{document}

%% file: appendix.tex
\begin{appendix}

{\Large \bf Appendix}
\appendix

\section{Details in creating $\delta$ for input independence}
The additional regularization terms in the loss function for creating $\delta$ for input independence test is as follows:
$$ \pazocal{R} = \eta_2\big[(x + \delta - p_{h})^+ + (p_{l} - x - \delta)^+\big] + \eta_3\sum{\delta \odot (J - I_c)}$$
$\eta_2\big[(x + \delta - p_{h})^+ + (p_{l} - x - \delta)^+\big]$ penalizes pixel values in $x+\delta$ that fall outside the valid pixel range $[p_{\textit{l}}, p_{\textit{h}}]$ (e.g., $[0, 255]$). The additional term $\eta_3\sum{\delta \odot (J - I_c)}$ minimizes updates to regions outside of the $\delta$ region represented by mask $I_c$ ($J$ is a matrix of ones). The overall loss function is:
$$L = \|f(x+\delta)-f(x)\|_2 - \eta_1\|\delta\|_2 + \pazocal{R}$$
The update rule for $\delta$ is:
$$\delta_{t+1} = \delta_t - \lambda \frac{\partial L}{\partial \delta}$$
where $\lambda$ is the step size (defaults to $500$). $\delta_0$ is initialized from a dog image to obtain solutions that are semantically meaningful.

\section{Discussions on $\delta$ versus common feature}
Note that there is a subtle difference between $\delta$ generated from the optimization procedure above versus the common feature (\CF) obtained from training. Intuitively, to find $\delta$, we are moving an input in the direction perpendicular to $\nabla_xf(x)$, but the gradient itself is fixed because the model is fixed. When training a model with \CF, $\nabla_xf(x)$ becomes small with respect to the \CF. This explains why we expect minimal attribution change when $\delta$ is added to the input during input independence testing, but expect small attribution to dog \CF~during input dependence testing.

\section{Other measures for input independence}
\begin{figure}[ht]
  \includegraphics[width=.8\linewidth]{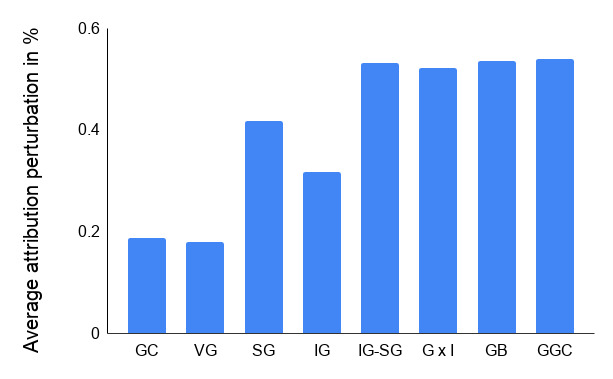}
  \setlength\belowcaptionskip{2ex}
  \caption{Average $g_c$ between $x$ and $x+\delta$ over 100 inputs.}
  \label{fig:ii_bar_raw}
\end{figure}

An alternative measure of input independence is the average attribution difference when $\delta$ is added to the input:
$$\frac{1}{|X_{corr}|}\sum\limits_{x \in X_{corr}}|\frac{g_c(f, x + \delta) - g_c(f, x)}{g_c(f, x)}| $$
Figure \ref{fig:ii_bar_raw} shows the average perturbation over 100 images for each saliency method. Lower attribution difference is better. The ranking is roughly the same as the \IIR~metric.

\section{DNN Architecture and training}
All BAM models are ResNet50~\cite{he2016deep} models. Training starts from an ImageNet pretrained checkpoint~\cite{Russakovsky15} (available at \url{https://github.com/tensorflow/models/tree/master/official/resnet}) and all layers are fine-tuned on the BAM dataset. 90\% of randomly chosen members of the BAM dataset are used to train, the rest are used for testing. All models are implemented using TensorFlow~\cite{abadi2016tensorflow} and trained on a single Nvidia Tesla V100 GPU. 

\section{Details of interpretability methods compared}

We consider neural network models with an input $x \in \mathbb{R}^d$ and a function $f(x): \mathbb{R}^d \mapsto \mathbb{R}^m$. A saliency method $e(f, x): \mathbb{R}^d \mapsto \mathbb{R}^d$ outputs a saliency map highlighting regions relevant to prediction. Below is an overview of the eight saliency methods evaluated in our work.

\textbf{GradCAM (GC)}~\cite{Selvaraju16} computes the gradient of the class logit with respect to the feature map of the last convolutional layer of a DNN. Guided GradCAM (GGC) is GC combined with Guided Backprop through an element-wise product.

\textbf{Vanilla Gradient (VG)}~\cite{Simonyan13,Erhan09,Baehrens10} computes the gradient of the target class $k$ at logit layer $l$ with respect to each input pixel: $e(f,x)=\frac{\partial f_l^k}{\partial x}$, reflecting how much the logit layer output would change when the input changes in a small neighborhood.

\textbf{SmoothGrad (SG)}~\cite{Smilkov17} reduces visual noise by averaging the explanations over a set of noisy images in the neighborhood of the original input image: $\frac{1}{|N|}\sum_{i=0}^N{e(f, x+z_i)}$, where $z_i \sim \mathcal{N}(\mu,\,\sigma^{2})$.

\textbf{Integrated Gradient (IG)}~\cite{Sundararajan17} computes the sum of gradients for a scaled set of images along the path between a baseline image ($x^\prime$) and the original input image: $e(f,x) = (x-x^\prime) \times \int_{0}^{1} \frac{\partial f(x^\prime + \alpha (x - x^\prime)}{\partial x}d\alpha$. Smoothing from SG can be applied to IG to produce IG-SG.

\textbf{Gradient x Input (G x I)} computes an element-wise product between VG and the original input image.~\cite{Ancona18} showed that for a ReLU network with zero baseline and no bias.

\textbf{Guided Backpropagation (GB)}~\cite{Springenberg14} builds on top of the DeConvNet explanation method~\cite{zeiler2014visualizing} and attributes input importance through backpropagating neuron activations from the logit layer to the input layer.

We accompany visualization of a subset of saliency methods by averaging over channels and capping the extremes to the $99^{th}$ percentile as done by~\cite{Sundararajan17,Smilkov17} before normalizing each attribution to between $[0,1]$. 

\section{Details of computing TCAV scores}
We compute the TCAV scores of the dog concept for different models (e.g. $f_o$ and $f_s$ for absolute contrast). To learn the dog CAV, we take 100 images from $X_{o,s}$ where the object is a dog as positive examples and 100 images from $X_{\varnothing,s}$ as negative examples for the dog concept. We perform two-sided t-test of the TCAV scores, and reject scores where $p$-value $>$ 0.01. We compute TCAV scores for each of the block layer and the logit layer of ResNet50. The final TCAV score is the average of the layers that passed statistical testing.

\section{Details of model contrast score baseline}
For model contrast score, we generate a random mask $I_c$ to calculate baseline differences. For TCAV, we obtain two TCAV scores for two models ($f_o$ and $f_s$), and show the difference between the two, both for TCAV scores for the dog CAVs and random CAVs. 

\section{Full size figures}

\begin{figure*}
  \includegraphics[width=1.\linewidth]{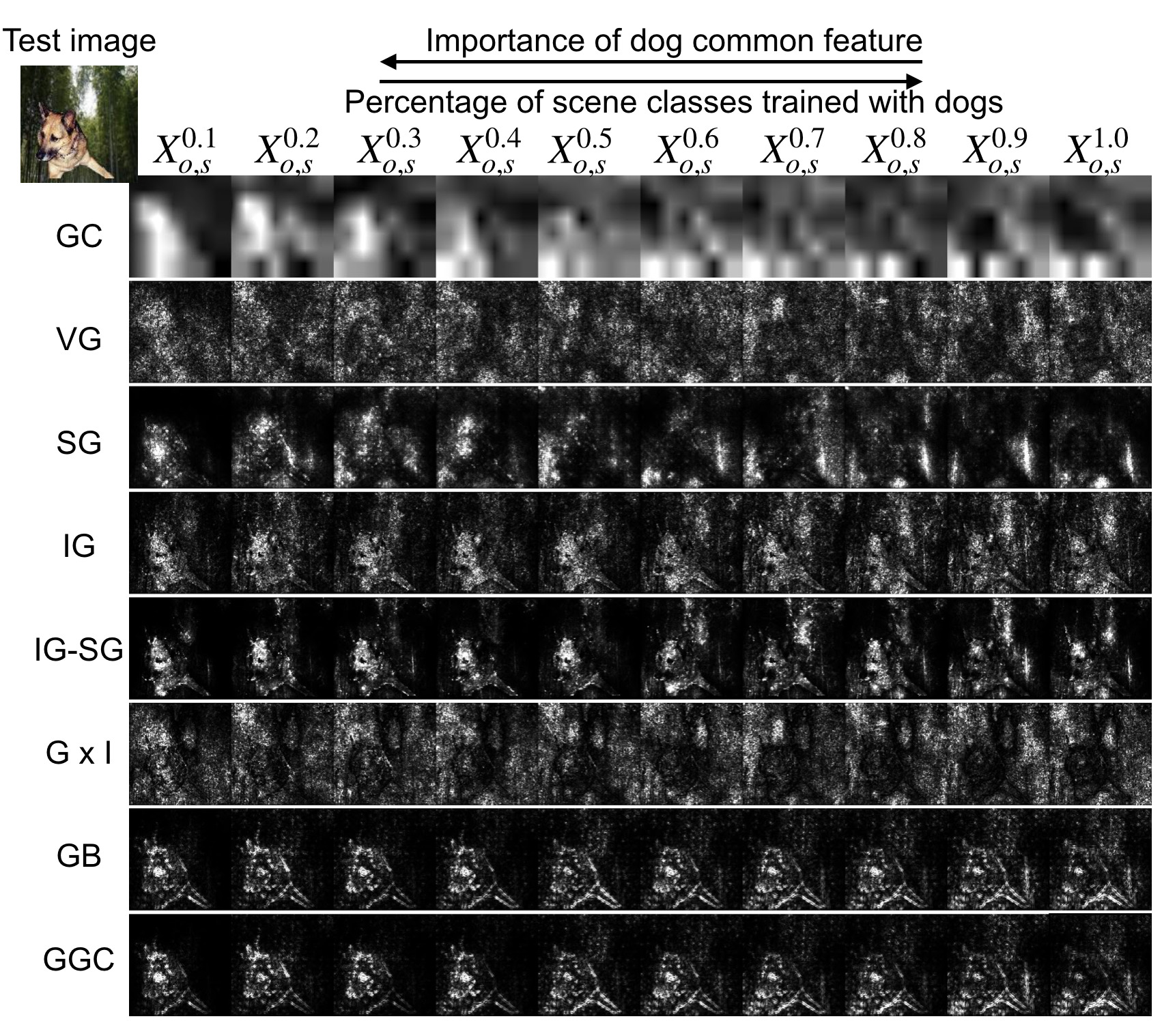}
  \caption{Full figure of an example of saliency map visualizations for models trained with \CF s of different $k$. $k$ increases from left to right. A larger contrast among each row is better.}
\end{figure*}
\begin{figure*}
  \includegraphics[width=1.\linewidth]{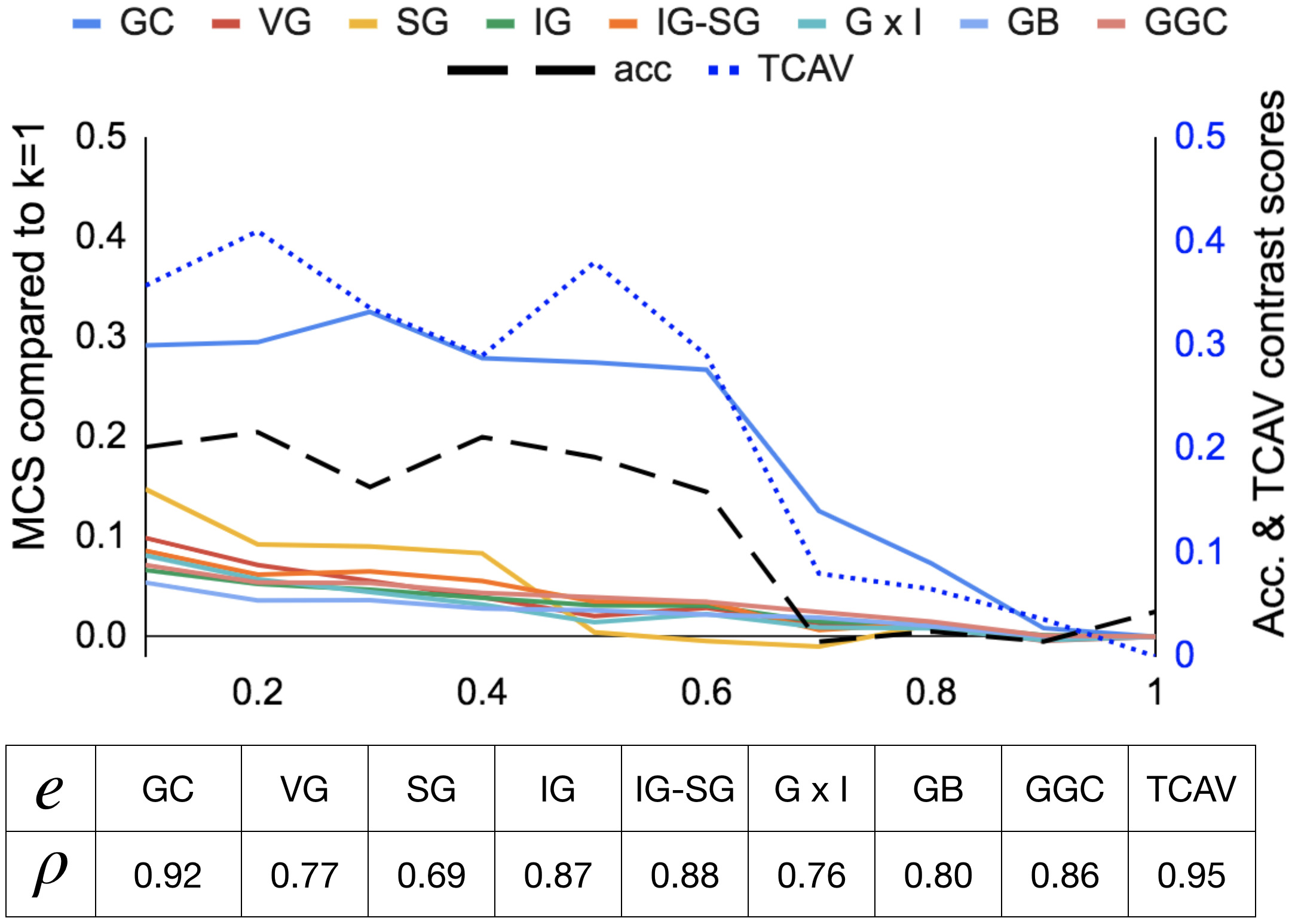}
  \caption{[Top] \MCS~between $\{X_{o,s}^k\}$ for $k \in \{0.1, \dots, 1.0\}$ and $X_{o,s}^{1.0}$ as $k$ increases. The dashed black line is the accuracy drop when \CF~is removed during testing. The dotted blue line is the series of \MCS~for TCAV. [Bottom] Pearson correlation coefficients ($\rho$) between each method's \MCS~and the accuracy drop. A higher correlation is better.}
\end{figure*}

\clearpage

\begin{figure*}[!]
  \includegraphics[width=1.\linewidth]{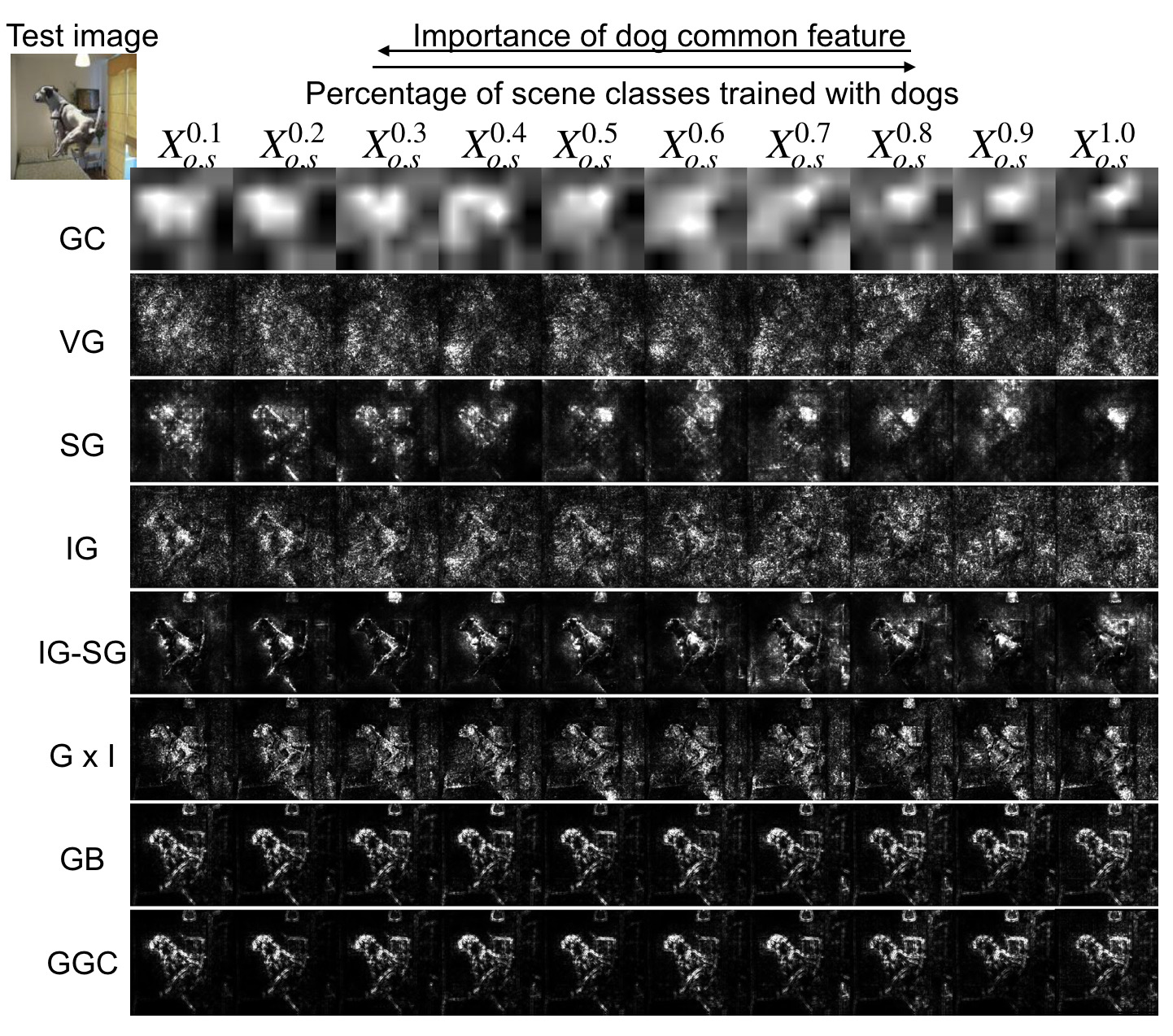}
    \caption{Additional example saliency maps from relative model contrast testing.}
\end{figure*}
\begin{figure*}
  \includegraphics[width=1.\linewidth]{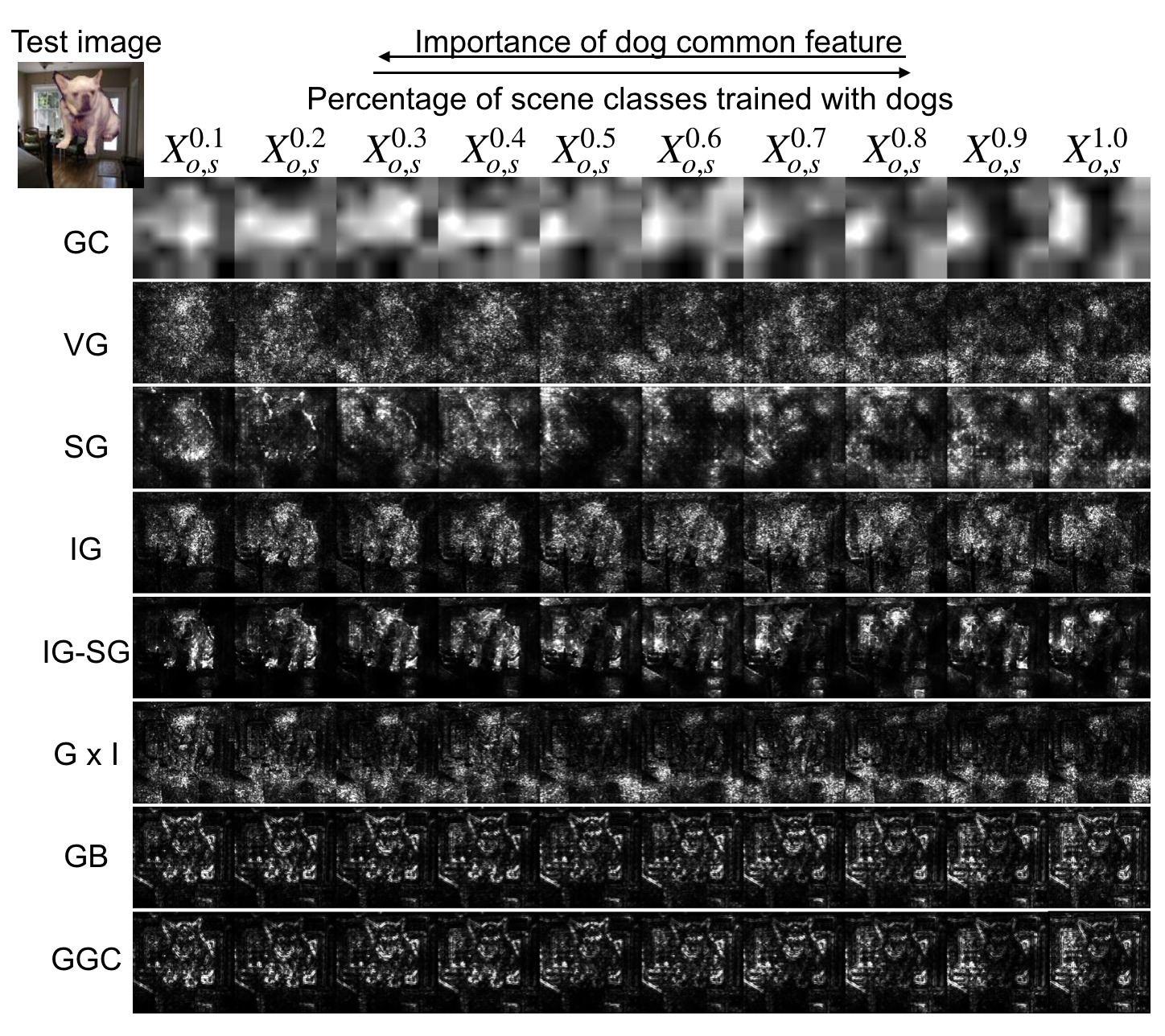}
  \caption{Additional example saliency maps from relative model contrast testing.}
\end{figure*}

\clearpage

\begin{figure*}
  \includegraphics[width=1.\linewidth]{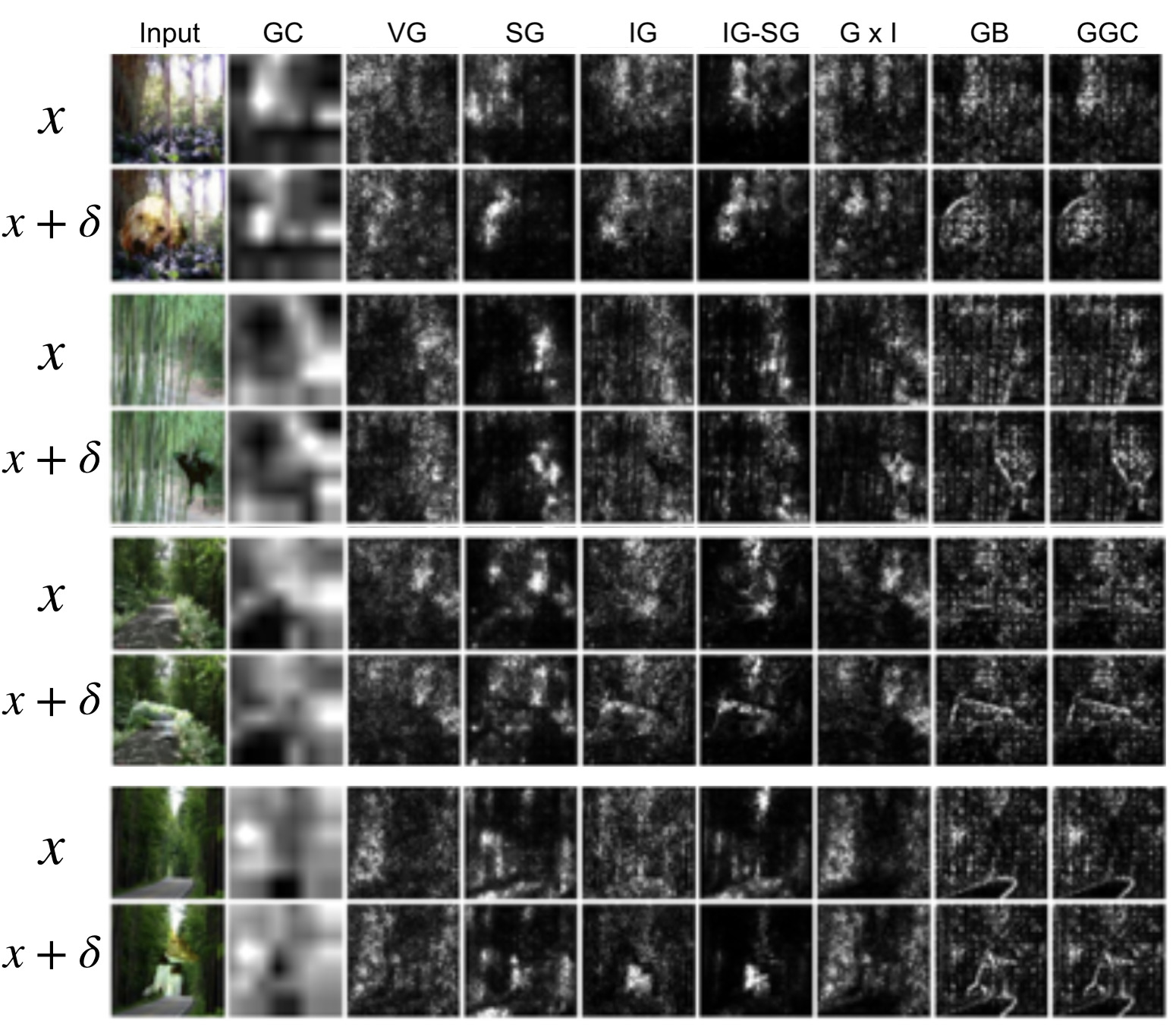}
  \caption{Additional example saliency maps from input independence testing.}
\end{figure*}

\end{appendix}